\documentclass[11pt]{article}

\usepackage[preprint]{acl}

\usepackage{url}
\usepackage{times}
\usepackage{latexsym}
\usepackage{hyperref}
\usepackage{lipsum}

\usepackage[T1]{fontenc}

\usepackage[utf8]{inputenc}

\usepackage{microtype}

\usepackage{inconsolata}

\usepackage{graphicx}
\usepackage{subcaption}
\usepackage{amsmath}
\usepackage{tikz}
\usetikzlibrary{external, shapes, arrows.meta, positioning, fit, backgrounds, calc, shadows}

\title{Listen, Attend, Understand: a Regularization Technique for Stable E2E Speech Translation Training on High Variance labels}

\author{
  Yacouba Diarra \and Michael Leventhal \\ 
  RobotsMali AI4D Lab, Bamako, Mali \\
  \texttt{diarray} \and \texttt{mleventhal} | \texttt{@robotsmali.org} 
}

\begin{document}
\maketitle
\begin{abstract}
End-to-End Speech Translation often shows slower convergence and worse performance when target transcriptions exhibit high variance and semantic ambiguity. We propose Listen, Attend, Understand (LAU), a semantic regularization technique that constrains the acoustic encoder's latent space during training. By leveraging frozen text embeddings to provide a directional auxiliary loss, LAU injects linguistic groundedness into the acoustic representation without increasing inference cost. We evaluate our method on a Bambara-to-French dataset with 30 hours of Bambara speech translated by non-professionals. Experimental results demonstrate that LAU models achieve comparable performance by standard metrics compared to an E2E-ST system pretrained with 100\% more data and while performing better in preserving semantic meaning. Furthermore, we introduce Total Parameter Drift as a metric to quantify the structural impact of regularization to demonstrate that semantic constraints actively reorganize the encoder's weights to prioritize meaning over literal phonetics. Our findings suggest that LAU is a robust alternative to post-hoc rescoring and a valuable addition to E2E-ST training, especially when training data is scarce and/or noisy.
\end{abstract}

\section{Introduction}
\label{sec:intro}
Following the success of Listen, Attend and Spell (LAS) \cite{chan2015listenattendspell}, End-to-End neural network architectures that simultaneously learn all the components of a speech recognizer have become the dominant paradigm in automatic speech recognition, reaching human level performance on a number of benchmarks \cite{humanparityASR}. However, even with near human-level performance on some benchmarks, ASR systems still produce non-negligible errors. For some especially sensitive applications, it has become standard to use an external pretrained language model for second-pass rescoring or post-hoc corrections of an ASR model's n-best candidate transcriptions, often referred to as hypotheses (\citealp{Ogawa_rescoring}; \citealp{futami2020distillingknowledgebertsequencetosequence}; \citealp{zhao2021bart}; \citealp{chen2023hyporadise}; \citealp{li2024investigating}). Cascading modern large language models with ASR can improve transcription quality by correcting the semantic and grammatical errors in first-pass ASR hypotheses, albeit with the resource costs of running a large LLM for second-pass correction.

Training-time robustness methods such as SpecAugment \cite{Park_2019} address the acoustic side of the problem. Less work targets robustness to ambiguity in transcriptions and translations, with research having gravitated toward post-hoc correction at inference time. Label ambiguity can slow convergence and reduce training stability. It is particularly pronounced in speech translation, where even with context, a source-language word or phrase often admits multiple acceptable translations \cite{Prior2009TranslationAI}. This issue is often amplified in low-resource languages (LRLs), since limited data and fewer standardized annotation conventions increase variability across annotators and domains, and smaller corpora provide fewer examples for the model to average out this noise.

In this paper, we introduce a new regularization technique (\textbf{LAU: Listen Attend Understand}) in the training of E2E Speech Translation models (E2E-ST) that aims to steer the acoustic encoder towards outputting representations that also exist in a high-resource semantic space defined by a pretrained embedding model rather than the acoustic space alone. Our experiments on Bambara-French speech translation with jeli-asr, a 30-hour semi-professionally translated dataset \cite{Diarra2022Griots}, show that this new training constraint is active with noticeable differences in the \textit{total parameter drift}, i.e., the L2 norm of the difference between the initial and final encoder weights, and outperforms a conventional E2E-ST on semantically-oriented tasks, despite significantly less pretraining data. 

We make the following two main contributions:

\begin{itemize}
    \item We show that, for Bambara-to-French speech translation, a simple encoder–decoder E2E-ST model trained on 63 hours of Bambara speech (with French translations available for only 30 hours) outperforms a cascaded ASR-to-machine-translation pipeline. This result offers evidence pertinent for LRL settings that avoiding the compounding of errors across two low-resource components can be beneficial, even though direct E2E is typically more difficult to learn.    
    \item We implement a new constraint for E2E-ST that stabilizes training on high variance non-professional translations through transfer learning, forcing the encoder's output to carry meaningful information in an external pretrained embedding space. We run experiments with the same 30-hour dataset and discuss the results in \ref{sec:results}.
\end{itemize}

\section{Related Works}
\label{sec:literature}
\subsection{End-to-End Speech Translation}
\label{subsec:e2e-st}
Given the triplet training corpus \texttt{\textless speech, transcription, translation\textgreater}, the conventional training method for high-quality E2E-ST involves pretraining an ASR baseline with the \texttt{\textless speech, transcription\textgreater} pair and
then optimizing it further using the \texttt{\textless speech, translation\textgreater} pair. However, most recent breakthroughs in E2E-ST have come from a dual-task approach where a neural network model is trained to perform both ASR and ST, either through multi-task prompt tokens with transformer-based architectures (\citealp{radford2022robustspeechrecognitionlargescale}; \citealp{puvvada2024moreaccuratespeechrecognition}) or jointly training two decoders, one for each task (\citealp{ko2021asr}; \citealp{chuang-etal-2020-worse}).

Previous work has shown that, in high-resource settings, direct E2E-ST can outperform a conventional ASR-to-MT cascaded pipeline when sufficient data is available (\citealp{weiss2017sequencetosequencemodelsdirectlytranslate}; \citealp{sperber-etal-2019-attention}). In Sections \ref{subsec:bleu_wer_resulys} and \ref{subsec:slu_results}, we show that the same qualitative trend can hold for a low-resource language such as Bambara, with as little as 30 hours of data. Beyond comparing end-to-end and cascaded paradigms, recent work has also investigated tighter integration between speech models and large language models by learning a trainable connector module (e.g. Dense, Transformer, or Q-former) between a speech encoder and an LLM, allowing the LLM to take speech-derived representations as input (\citealp{sedláček2024aligningpretrainedmodelsspoken}; \citealp{yu2023connectingspeechencoderlarge}).

For more robust E2E-ST training, \citeauthor{ko2021asr} also added ASR confusion as a factor in multi-task training. They introduce a posterior-based ASR loss in which target tokens are replaced with the posterior distribution produced by a pretrained ASR model. This softer reference encodes competing hypotheses and injects ASR confusion into the learning signal. Since the ST decoder takes both the ASR decoder and the encoder's states as input, this uncertainty-aware ASR constraint helps the translation model cope with recognition ambiguity. The approach demonstrated better BLEU results in their Fisher Spanish-to-English data \cite{ko2021asr}.

\citeauthor{du2022regularizingendtoendspeechtranslation} argue that the conventional E2E-ST training method fails to exploit the association between triplet data by decoupling them into two-tuple data (pairs) at each stage. They train a single encoder-decoder model to learn the joint probability of transcription and translation with a dual-path successive decoding method. The decoder outputs the concatenated log probabilities of the two tasks either in ASR-MT order or, inversely, in ST-BT (Back Translation) order. They also introduce two Kullback-Leibler divergence regularization terms into the model training objective to reduce the mismatch between the output probabilities of the two paths. Their approach demonstrated gains in both ASR and ST tasks on the MuST-C benchmark over the conventional decoupled approach \cite{du2022regularizingendtoendspeechtranslation}.

\subsection{Introducing Semantics in Spoken Language Understanding}
\label{subsec:semantic-in-slu}
Recent work has highlighted that self-supervised learning (SSL) speech encoders (\citealp{chung2019unsupervisedautoregressivemodelspeech}; \citealp{baevski2020wav2vec20frameworkselfsupervised}), which are widely used as general-purpose front ends for spoken language understanding tasks such as ASR, intent classification, and question answering, tend to emphasize acoustic and phonetic structure, with comparatively weaker semantic information in their representations \cite{pasad2022layerwiseanalysisselfsupervisedspeech}.

In this vein, \citeauthor{xu-etal-2023-introducing} augment SSL speech encoders with semantic information from an LLM by using unsupervised ASR (ASR-U) as a bridge that converts speech-encoder embeddings into the subword token inputs used by the LLM. Their bridge follows wav2vec-U 2.0 by using an adversarial (GAN) component to predict phoneme sequences from SSL features, and then applies a WFST decoder constructed from a lexicon that encodes known phoneme-to-subword mappings to convert the predicted phonemes into the LLM’s subword tokens. While they achieve impressive results across multiple SLU tasks without labeled ASR data \cite{xu-etal-2023-introducing}, their approach is complex, introducing an additional ASR-U bridge and their focus was on the universal representation framework. By contrast, our LAU implementation is task-directed, adding a single auxiliary constraint tailored to stabilizing E2E-ST for a specific language pair, rather than optimizing a universal representation intended to transfer broadly across tasks.

\citeauthor{chuang-etal-2020-worse} propose a semantically aware E2E speech translation by encouraging the ASR-decoder hidden states to align with pretrained source-language word embeddings in a dual-decoder setting. In our implementation, we use a single-decoder architecture and apply the embedding-alignment constraint at the encoder level using a training-only semantic head that projects encoder outputs into the same dimensionality as the embedding model (target-language in our case). This constrains the most information-rich representation of the audio input before it is mapped to a limited output vocabulary. At inference time, the semantic head is removed, so the regularizer introduces no additional cost.

In high-dimensional embedding spaces, cosine similarity is a common choice for implementing such alignment because it measures the angle between vectors, which often correlates with semantic relatedness, and because many embedding models are trained with objectives that promote directional agreement (\citealp{mikolov2013efficientestimationwordrepresentations}; \citealp{grill_byol}; \citealp{chen2020simpleframeworkcontrastivelearning}).

Finally, while cosine similarity is widely used because it is scale-invariant and matches the training objectives of many embedding models, recent work suggests that embedding magnitude can also encode meaningful semantic signals that cosine discards \cite{you2025semanticsanglecosinesimilarity}. Motivated by this observation, we consider both cosine-based and magnitude-sensitive alignment losses when instantiating our encoder-level embedding-alignment regularizer.

\section{Implementation \& Experiments}
\label{sec:impl}
We used \texttt{RobotsMali/soloni-114m-tdt-ctc-v0} (soloni-v0), a Bambara finetune of NVIDIA's 114-M parameter Parakeet model (\citealp{rekesh2023fastconformerlinearlyscalable}; \citealp{xu2023efficientsequencetransductionjointly}) trained on jeli-asr \cite{Diarra2022Griots} and \texttt{RobotsMali/soloni-114m-tdt-ctc-v1} (soloni-v1), finetuned over this model on Kunkado \cite{diarra2025kunnafonidilawkacadeauasr} as our ASR baselines. 

We further trained soloni-v1 on the \texttt{\textless speech, translation\textgreater} pairs of jeli-asr for $12,000$ steps, using the AdamW optimizer and Noam scheduler with a learning rate scaling factor of $2.0$ and 1k warmup steps \cite{vaswani2023attentionneed}, for use as our traditional E2E-ST model.

We use soloni-v0 as baseline for all our LAU experiments and \texttt{dangvantuan/sentence-camembert-base} as our pretrained French embedding model (\citealp{reimers2019sentence}; \citealp{martin2020camembert}); we then added a shallow semantic head with only two \textit{fully-connected layers} to map the encoder's output to the same dimension as our pretrained embeddings. We wanted this semantic head to be lightweight so that the semantic loss has greater impact on the encoder. We trained the updated architecture with the weighted sum of the loss of the sequence-to-sequence task (ASR/ST) and the semantic loss

\begin{equation}
    \mathcal{L}_{\text{LAU}} = \mathcal{L}_{\text{Seq}} + \lambda \times \mathcal{L}_{\text{semantic}}
    \label{equa:lau}
\end{equation}
where $\lambda$ is a hyperparameter controlling the strength of the regularization.

\paragraph{Sequence Loss.} This term is the loss function for the primary seq2seq task that we are trying to learn (ST in this experiment), usually a transducer loss. In our case, since the Parakeet model has a hybrid architecture with two independent decoders (TDT \& CTC), it is also a sum of losses:

\begin{equation}
    \mathcal{L}_{\text{Seq}} = (1 - \alpha) \times \mathcal{L}_{\text{TDT}} + \alpha \times \mathcal{L}_{\text{CTC}}
    \label{equa:seq}
\end{equation}

where $\alpha$ is also a hyperparameter; with:
\begin{equation}
    \mathcal{L}_{\text{TDT}} = -\log P_{\text{TDT}}(\mathbf{y}|\mathbf{x})
    \label{equa:tdt}
\end{equation}
where $P_{\text{TDT}}(\mathbf{y}|\mathbf{x})$ is the total probability of the target sequence $\mathbf{y}$, calculated by marginalizing over all valid joint alignments of tokens (including blanks) and durations \cite{xu2023efficientsequencetransductionjointly}; and:

\begin{equation}
\mathcal{L}_{\text{CTC}} = -\log \sum_{\pi \in \mathcal{B}^{-1}(\mathbf{y})} P(\pi|\mathbf{x})
\label{equa:ctc}
\end{equation}
where $\mathcal{B}^{-1}(\mathbf{y})$ represents the set of all possible frame-level paths $\pi$ that map to the target sequence $\mathbf{y}$ after removing blanks and repetitions \cite{gravesctc2006}.

Our implementation is based on NVIDIA's NeMo toolkit which already implements these training objectives \cite{kuchaiev2019nemotoolkitbuildingai}. We made minimal modifications to the open toolkit source code to integrate the embedding model and the semantic loss.

\paragraph{Semantic Loss.} For the semantic regularization term, we experimented with two classic similarity measures, the cosine embedding loss (\ref{equa:cosine_loss}) and the MSE loss (\ref{equa:mse}).

\begin{equation}
    \mathcal{L}_{\text{cosine}} = 1 - \frac{Y \cdot \hat{Y}}{\|Y\| \|\hat{Y}\|}   
    \label{equa:cosine_loss}
\end{equation}

\begin{equation}
    \mathcal{L}_{\text{MSE}} = \frac{1}{N} \sum_{i=1}^N (Y_i - \hat{Y}_i)^2
    \label{equa:mse}
\end{equation}
where $\hat{Y}$ is the output of the semantic head and $Y$ is the reference embedding vector obtained by encoding the reference translation with our French embedding model.

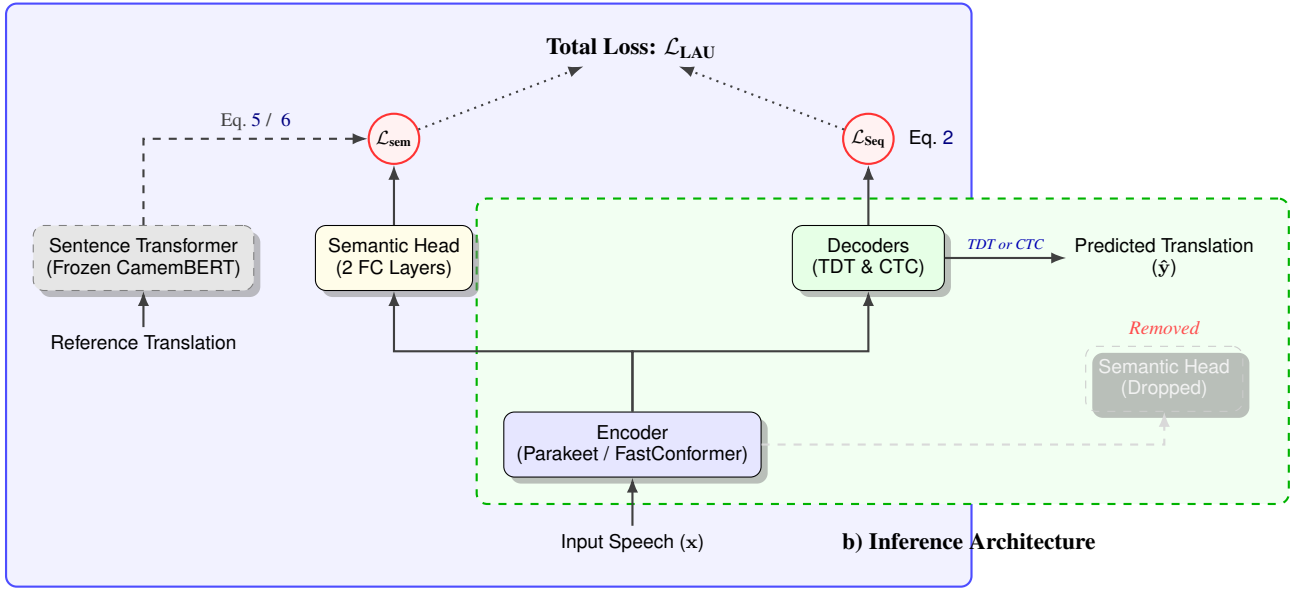
\begin{figure*}[htbp]
    \centering
    \begin{tikzpicture}[scale=0.8, transform shape, node distance=1cm]

        \tikzset{
            base/.style = {
                draw, rounded corners, align=center,
                minimum height=2.8em,
                drop shadow,
                font=\small\sffamily,
                inner sep=6pt
            },
            encoder/.style = {base, fill=blue!10, minimum width=3.2cm},
            decoder/.style = {base, fill=green!10, minimum width=2.5cm},
            frozen/.style  = {base, fill=gray!20, dashed, draw=gray},
            loss/.style = {
                circle, draw=red!80, thick,
                fill=red!5, inner sep=2pt,
                font=\bfseries\small
            },
            conn/.style = {-{Latex[length=2mm]}, thick, darkgray},
            txt/.style = {font=\footnotesize\sffamily},
            block/.style = {
                draw, thick, rounded corners, inner sep=10pt
            }
        }

        \node[txt] (input) {Input Speech ($\mathbf{x}$)};
        \node[encoder, above=0.8cm of input] (enc)
            {Encoder\\(Parakeet / FastConformer)};
        \draw[conn] (input) -- (enc);

        \node[decoder, above right=2cm and 0.5cm of enc] (dec)
            {Decoders\\(TDT \& CTC)};
        \draw[conn] (enc.north) -- ++(0,1.0) -| (dec.south);

        \node[base, fill=yellow!10,
              above left=2cm and 0.5cm of enc] (semhead)
            {Semantic Head\\(2 FC Layers)};
        \draw[conn] (enc.north) -- ++(0,1.0) -| (semhead.south);

        \node[frozen, left=1cm of semhead] (refmodel)
            {Sentence Transformer\\(Frozen CamemBERT)};
        \node[txt, below=0.6cm of refmodel] (refinput)
            {Reference Translation};
        \draw[conn] (refinput) -- (refmodel);

        \node[loss, above=1cm of semhead] (lsem)
            {$\mathcal{L}_{\text{sem}}$};
        \draw[conn] (semhead) -- (lsem);
        \draw[conn, dashed] (refmodel.north) |- (lsem.west)
            node[pos=0.75, above, font=\footnotesize] {Eq.~\ref{equa:cosine_loss} / ~\ref{equa:mse}};

        \node[loss, above=1cm of dec] (lseq)
            {$\mathcal{L}_{\text{Seq}}$};
        \node[txt, right=0.1cm of lseq] {Eq.~\ref{equa:seq}};
        \draw[conn] (dec) -- (lseq);

        \coordinate (combine) at ($(lseq)!0.5!(lsem)$);
        \node[txt, above=1.2cm of combine, font=\bfseries] (total)
            {Total Loss: $\mathcal{L}_{\text{LAU}}$};
        \draw[conn, dotted] (lsem) -- (total);
        \draw[conn, dotted] (lseq) -- (total);

        \node[txt, right=2cm of dec, align=center] (output)
            {Predicted Translation\\($\mathbf{\hat{y}}$)};
        \draw[conn] (dec.east) -- (output.west)
            node[midway, above, font=\scriptsize\itshape, text=blue!70!black]
            {TDT or CTC};

        \node[base, dashed, draw=gray!40, text=gray!10,
              below=1cm of output] (ghost)
            {Semantic Head\\(Dropped)};
        \node[font=\footnotesize\itshape, text=red!70,
              above=0.05cm of ghost] {Removed};
        \draw[conn, dashed, gray!30] (enc.east) -| (ghost.south);

        \begin{scope}[on background layer]

            \node[block, draw=blue!70, fill=blue!5]
                (trainbox)
                [fit=(refmodel)(semhead)(lsem)(lseq)(total)
                     (input)(enc)(dec)] {};

            \node[below=0.5cm of trainbox, font=\bfseries]
                {a) Proposed Training Architecture};

            \node[block, draw=green!70!black, dashed, fill=green!5]
                (inferbox)
                [fit=(enc)(dec)(output)] {};

            \node[below=0.5cm of inferbox, right=2cm of input, font=\bfseries]
                {b) Inference Architecture};

        \end{scope}

    \end{tikzpicture}

    \caption{\textbf{a)} The training graph adds a shallow semantic head to the encoder output to align it with frozen reference embeddings (LAU regularization). \textbf{b)} During inference, the semantic branch is dropped, resulting in a standard Encoder-Decoder architecture.}
    \label{fig:train_vs_infer_graphs}
\end{figure*}

During training, the $\text{SentenceTransformer}$ model is kept frozen and only provides reference embeddings for the computation of the semantic loss. The gradients of the semantic loss only flow into the shallow semantic head and the encoder, thereby leveraging transfer learning from the pretrained embedding space. The speech translation part is independently trained. At inference time, the semantic head is removed, so the inference graph is identical to the baseline model, and decoding can be performed using either the CTC or the TDT decoder. Figure \ref{fig:train_vs_infer_graphs} illustrates the training and inference graphs.

\begin{figure}[h!]
    \centering
    \includegraphics[width=\linewidth]{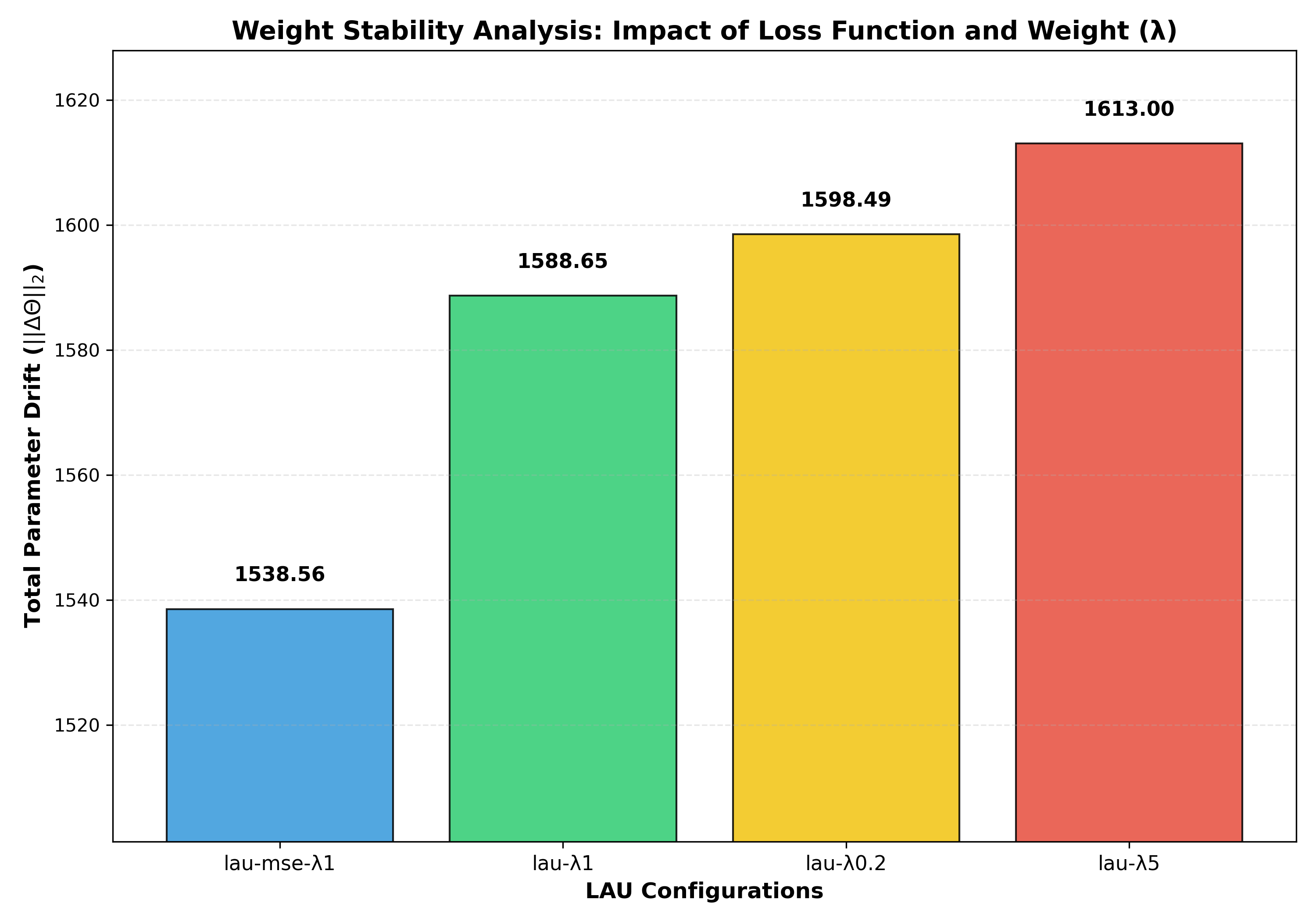}
    \caption{Total parameter drift for different semantic regularization loss choices and weightings}
    \label{fig:param_drift}
\end{figure}

\section{Results}
\label{sec:results}
We ran training experiments with the proposed architecture and test with both cosine and MSE as semantic loss with no weight as a baseline ($\lambda = 1.0$). We also explored the effect of regularization with different semantic loss weights ($\lambda = 0.2$ and $\lambda = 5.0$), however, due to compute resource limitations we only ran experiments with different weightings for the cosine loss (the observations likely generalize to the MSE loss as well). All these experiments ran for $11,800$ steps over soloni-v0 with the same optimizer configuration as for our conventional E2E-ST model.

\subsection{LAU for Regularization}
\label{subsec:reg_results}
We evaluate the effectiveness of LAU regularization by \textit{total parameter drift}, that is, comparing the L2 norm of the difference between the encoder weights before and after training (\ref{equa:drift}). This metric indicates how strongly the regularization loss constrains changes to the encoder: smaller values correspond to less substantial updates, that is, stronger effective regularization. Our experiments suggest that the MSE loss is best suited for regularization and that the optimal value for $\lambda$ is around $1.0$. Much lower values (e.g., 0.2) result in larger parameter changes, leading the model to overfit the high-variance translations, whereas much higher values cause the auxiliary semantic objective to dominate the main task, resulting in even more substantial updates, as shown in figure \ref{fig:param_drift}. This aligns with recent findings that cosine similarity may not be the best semantic similarity measure in certain contexts and its scale-invariance could cause it to miss important semantic information \cite{you2025semanticsanglecosinesimilarity}

\begin{equation}
    \text{Drift} = ||\Theta_{final} - \Theta_{initial}||_2
    \label{equa:drift}
\end{equation}
where $\Theta$ represents the weights of the encoder.

\subsection{BLEU \& WER evaluation}
\label{subsec:bleu_wer_resulys}
The Bilingual Evaluation Understudy (BLEU), the  Word Error Rate (WER) and the Character Error Rate (CER) are the standard metrics for evaluating speech translation systems. Table \ref{tab:wer_bleu_cer_summary} reports the BLEU-4, WER and CER scores achieved by our E2E-ST model (soloni-v1) trained on 30 hours of jeli-asr and 33 hours of Kunkado audio and the LAU-trained models (soloni-v0) trained on the 30 hours of of jeli-asr audio \cite{Diarra2022Griots}, and compare with the ASR$\rightarrow$MT pipeline using soloni-v0 and Djelia's machine translation model \cite{djelia2025}.

\begin{table}[ht]
    \centering
    \setlength{\tabcolsep}{4pt}
    \begin{tabular}{lccc}
        \hline
        \textbf{Model (Decoding)} & \textbf{WER $\downarrow$} & \textbf{CER $\downarrow$} & \textbf{BLEU $\uparrow$} \\ \hline
        ASR-MT (tdt)     & 0.9109 & 0.6884 & 0.0880 \\
        E2E-ST (tdt)        & \textbf{0.7043} & 0.5817 & \textbf{0.2418} \\
        LAU-$\lambda$0.2 (ctc)   & 0.7451 & 0.5742 & 0.1642 \\
        LAU-$\lambda$5 (ctc)     & 0.7455 & \textbf{0.5683} & 0.1611 \\
        LAU-$\lambda$1 (ctc)     & 0.7553 & 0.5858 & 0.1592 \\
        LAU-mse-$\lambda$1 (ctc) & 0.7608 & 0.5864 & 0.1429 \\ \hline
    \end{tabular}
    \caption{Comparison of model performance across WER, CER, and BLEU metrics. Since each model has two decoders, the decoding strategy yielding the best BLEU is reported. The values in bold represent the best scores per metric}
    \label{tab:wer_bleu_cer_summary}
\end{table}

Although the LAU models do not outperform the the fully pre-trained E2E-ST model, pretrained with 33 more hours of ASR data \cite{diarra2025kunnafonidilawkacadeauasr}, they get remarkably close, even surpassing it in CER ($-1.3\%$ approx.) for the LAU model trained with cosine loss and $\lambda = 5$.

Most notably, the fully pretrained E2E-ST model and all LAU-trained variants outperform the cascaded ASR-to-MT pipeline across all three metrics. This result extends prior findings from higher-resource settings \citealp{weiss2017sequencetosequencemodelsdirectlytranslate,sperber-etal-2019-attention} and supports the value of end-to-end training in resource-limited contexts, where error compounding across two low-resource components can be especially detrimental.

\subsection{LAU Potential for SLU}
\label{subsec:slu_results}
\begin{figure}[htbp]
     \centering
     \begin{subfigure}[b]{0.48\textwidth}
         \centering
         \includegraphics[width=\textwidth]{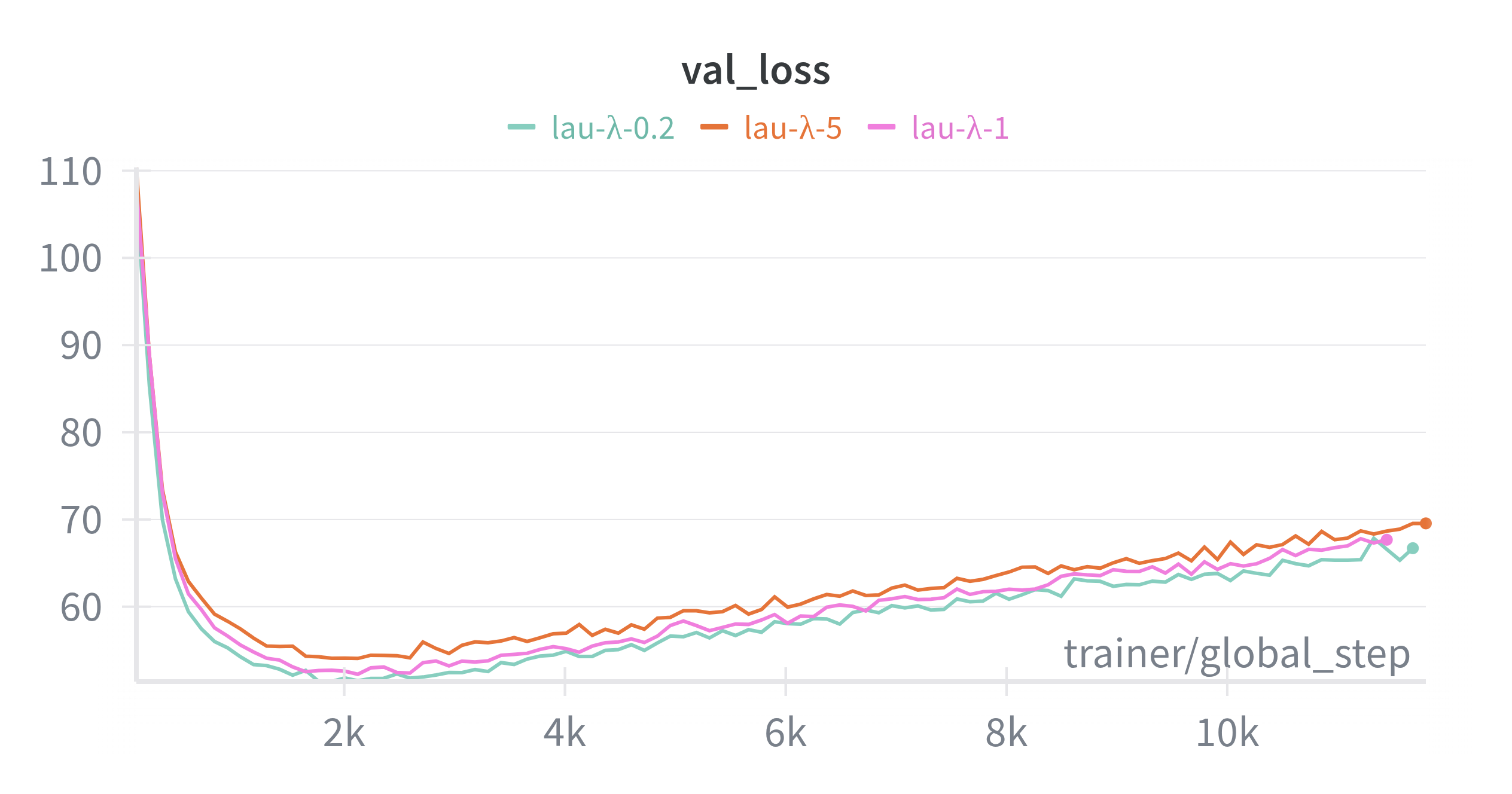}
         \caption{Total weighted loss sum (\ref{equa:lau})}
         \label{fig:total_val_loss}
     \end{subfigure}
     \hfill 
     \begin{subfigure}[b]{0.48\textwidth}
         \centering
         \includegraphics[width=\textwidth]{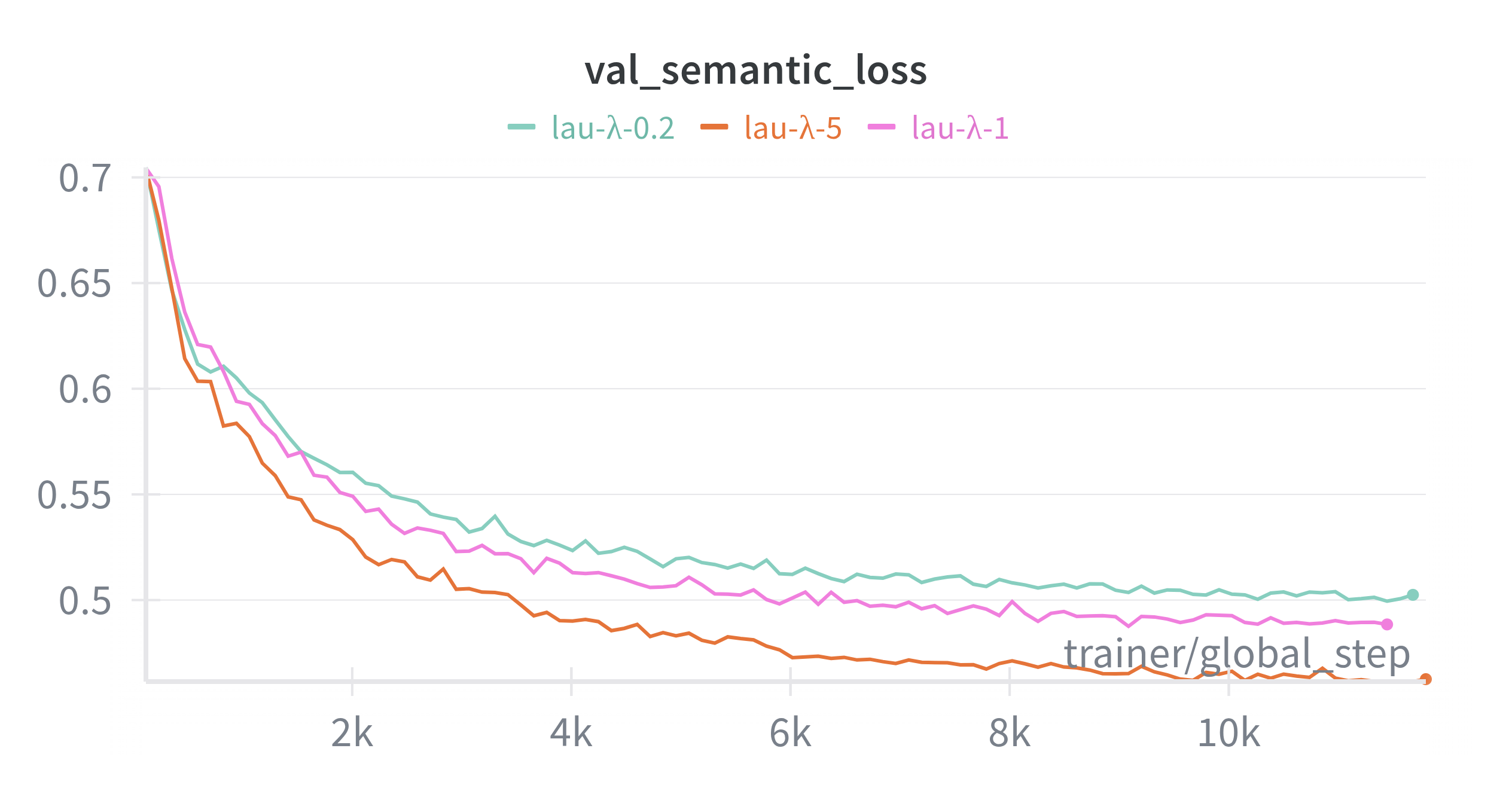}
         \caption{Semantic loss (\ref{equa:cosine_loss})}
         \label{fig:semantic_val_loss}
     \end{subfigure}
     
     \caption{Multitask conflict shows the new regularization technique is active as higher values for $\lambda$ drive more encoder updates to align with the semantic constraint to the detriment of lexical accuracy}
     \label{fig:multitask_conflict}
\end{figure}

In their multi-task implementation, \cite{chuang-etal-2020-worse} reported degradation in WER but better BLEU after integrating an embedding model. In our implementation, the embedding model is kept frozen and the semantic loss is applied only as an auxiliary regularizer, which appears to keep WER and BLEU more consistent with each other (Table \ref{tab:wer_bleu_cer_summary}). Nevertheless, we observe multi-objective interference: total validation loss decreases initially, but later increases while the semantic loss continues to decrease. This pattern suggests a genuine trade-off, indicating that the semantic loss is active and imposes a meaningful constraint on the shared encoder that the TDT/CTC objective must accommodate (see figure \ref{fig:multitask_conflict}).

\citealp{chuang-etal-2020-worse}'s findings also suggest that WER and BLEU may not be adequate metrics if we want to measure whether the output of the model has the correct semantics. With this in mind, we evaluate the models on two SLU-style tasks: LLM-based question answering (QA) and topic-based audio clustering (TAC). The QA task assesses whether a model’s outputs provide sufficient information for an LLM to answer questions or perform one-shot instructions based on the audio content, whereas TAC evaluates how well the outputs support grouping utterances by topic or intent.

For LLM-QA, we used a LLAMA-3.1-8B instruction-tuned model to generate 2230 French questions and answers from the audio utterances in our test set, and to assign a topic to each audio utterance based on the reference translation \cite{grattafiori2024llama3herdmodels}. We used the same model to answer the generated questions based only on the given model's output and, finally, we also used it in an LLM-as-a-judge pipeline to evaluate the correctness of the LLM answers \cite{gu2025surveyllmasajudge}.

It classified our test set into 6 main topic groups: family and origins, transmission of knowledge, society and culture, conflicts and rivalries, religion and spirituality, and "other topics". For the TAC task, we use k-means clustering to create 6 clusters in the same test data using only the outputs of the different models, by minimizing the cosine distance of their embeddings. Table \ref{tab:semantic_closeness_metrics} presents the LLM-QA accuracy (Acc), the cluster purity (Pur) and normalized mutual information (NMI) of the different models.

\paragraph{Cluster purity.} Cluster purity represents the average percentage of the "dominant" ground-truth topic within each k-means cluster. A purity of $1.0$ would mean that every cluster contains only one LLM-assigned topic.

\begin{equation}
    Purity = \frac{1}{N} \sum_{k=1}^{K} \max_{j} | \omega_k \cap c_j |
    \label{equa:purity}
\end{equation}

Where $N$: Total number of samples. $\omega_k$: The set of points in the $k$-th k-means cluster and $c_j$ the set of points in the $j$-th ground-truth class (from the LLM).

\paragraph{Normalized Mutual Information (NMI).} While purity measures homogeneity, NMI captures overall agreement by accounting for both homogeneity and completeness. In particular, if a semantic category (topic) is split across many clusters, NMI penalizes this fragmentation. It gives an idea of how closely the k-means clusters match the LLM-derived ones. NMI can reveal when the clustering model isolates just a few topics while failing to capture others, a situation purity alone may not identify.

\begin{equation}
    NMI(\Omega, C) = \frac{2 \times I(\Omega; C)}{H(\Omega) + H(C)}
\end{equation}

Where: $I(\Omega; C)$: Measures how much information is shared between the k-means clusters ($\Omega$) and the ground-truth classes ($C$). $H(\Omega)$ and $H(C)$ are the entropies of the k-means clusters and the LLM-assigned clusters, respectively.

\begin{table}[ht]
    \centering
    \setlength{\tabcolsep}{4pt}
    \begin{tabular}{lccc}
        \hline
        \textbf{Model (Decoding)} & \textbf{Acc $\uparrow$} & \textbf{Pur $\uparrow$} & \textbf{NMI $\uparrow$} \\ \hline
        ASR-MT (tdt)     & 0.3165 & 0.5947 & 0.0590 \\
        E2E-ST (tdt)        & 0.3762 & 0.6177 & 0.0543 \\
        LAU-$\lambda$0.2 (ctc)   & 0.3789 & 0.6184 & 0.0623 \\
        LAU-$\lambda$5 (ctc)     & 0.3721 & 0.6150 & 0.0592 \\
        LAU-$\lambda$1 (tdt)     & 0.3740 & \textbf{0.6279} & 0.0673 \\
        LAU-mse-$\lambda$1 (ctc) & \textbf{0.3834} & 0.6218 & \textbf{0.0705} \\ \hline
    \end{tabular}
    \caption{Comparison of model performance across LLM-QA accuracy, cluster purity, and normalized mutual information.}
    \label{tab:semantic_closeness_metrics}
\end{table}

This evaluation suggests that the LAU-trained models retain more semantic information, which is beneficial for downstream ST, particularly when combined with LLMs, and for other SLU tasks. Across all models, the “family and origins” topic is identified with very high accuracy (65–91\%). However, several of the evaluated models split this same topic across multiple clusters that are each dominated by “family and origins” examples, a behavior that is likely reinforced by the underlying class imbalance. As a result, purity remains relatively high, since clusters are internally homogeneous, but NMI stays low (6–7\%) because the clustering fails to provide a complete, one-to-one partition of topics. Finally, while the ASR–MT pipeline achieves a competitive NMI (5.9\%), it is substantially outperformed by the other models in both accuracy and purity\footnote{The jeli-asr \href{https://huggingface.co/datasets/RobotsMali/lau-eval}{test set} with the model-generated translations, the E2E-ST model and the LAU-mse model are publicly available on RobotsMali's hugginface profile; along with a link to all the codes and prompts used for training and evaluation}.

\section{Discussion}
\label{sec:discuss}

\subsection{High Potential for LRL Speech Applications}
\label{lau-for-app}
Bambara has a phonemic orthography, meaning there is a consistent grapheme-phoneme correspondence (\citealp{konta2014}; \citealp{vydrin:halshs-03909864}), which could make it ideal for speech recognition, except that other factors (unstable standards, illiteracy, primacy of oral over written expression) make developing annotated data for Bambara ASR especially arduous. While E2E-ST is a significantly more complex task than ASR, recent studies suggest that it could be easier to find Bambara speakers with a sufficient level of French literacy to directly translate Bambara speech into French than to find Bambara speakers able to accurately transcribe their mother tongue \cite{diarra2025costanalysishumancorrectedtranscription}. 

Our findings can contribute to the development of practical speech applications for many LRLs with similar literacy issues. Creating ST datasets may be cheaper and more effective data strategy and as most modern applications rely on LLMs that are already mature with high-resource languages, E2E-ST has promise as a faster, more practical path to integrating with those applications.

In this paper, we not only present a technique that can stabilize E2E-ST training on non-professionally translated datasets, but also demonstrate that, by incorporating a semantic constraint in the speech encoder, LAU outperforms conventional E2E-ST for LLM-oriented applications and speech understanding (\ref{subsec:slu_results}).

\subsection{An Unsupervised Approximation of ASR}
Topic-based audio clustering (TAC), although used here only for evaluation, can also support limited transcript-free semantic indexing for applications that would otherwise rely on ASR. If a database already contains a subset of items with topic or intent labels, these labeled examples can be used to name or anchor the clusters identified by TAC. New, unlabeled audio can then be assigned to the nearest cluster, enabling topic-level search and filtering without requiring word-level transcriptions.

Since translations provide a text representation of each utterance, this approach could also be extended to cross-lingual clustering by mapping speech into a high-resource language, where category labels can be derived more reliably. This is a promising direction because our LAU results suggest that translations that are semantically faithful, even if not lexically exact, can still preserve the information needed for effective unsupervised clustering.

\section{Conclusion}
\label{sec:conclude}
In this work, we introduced a novel regularization framework designed to stabilize E2E Speech Translation in the face of high-variance training labels. By enforcing semantic alignment between the acoustic encoder and a pre-trained semantic embedding space, we demonstrated that a model can achieve competitive lexical accuracy (CER) and capture more underlying semantic information (LLM-QA accuracy, NMI) even without large-scale pre-training. Furthermore, our analysis of the magnitude of updates to the parameters of the constrained speech encoders confirmed that the semantic loss is an active optimization force, with high weights ($\lambda=5$) forcing a meaningful structural reorganization of the encoder. Ultimately, LAU regularization offers a computationally efficient pathway for developing more robust ST systems, especially for LRLs.

\section*{Author Contributions}
\label{sec:contrib}
Yacouba Diarra conceived the research direction, designed and conducted the experiments, and wrote the manuscript.
Michael Leventhal supervised the project as Principal Investigator and contributed to editing the manuscript.

\section*{Limitations}
\label{sec:limits}
Despite the promising results, this study has several limitations. First, the effectiveness of the LAU regularization depends on the availability of a high-quality pretrained text embedding model in the target language (French). Second, while the occurrence of multitask conflict and the parameter drift analysis indicate that the mechanism is active, we have not yet established a theoretical upper bound for the regularization weight $\lambda$ beyond which ASR/ST performance may collapse. Finally, our evaluation is limited to a relatively small corpus and a constrained experimental budget, which restricted the breadth of model and hyperparameter exploration. At the same time, this setting may be beneficial in that it reflects the conditions under which many low-resource language efforts operate, where collecting parallel data and running large sweeps is often infeasible. Further validation on diverse language pairs and larger datasets is required to confirm the generalizability of the "semantic pull" to higher-resource languages. 

\bibliography{custom}

\end{document}